\documentclass[acmtog,nonacm]{acmart}

\author{Liangchen Li}
\email{liangchen@mail.ustc.edu.cn}
\orcid{0009-0004-8176-0745}
\affiliation{%
  \institution{University of Science and Technology of China}
  \country{China}
}

\author{Caoliwen Wang}
\email{wclw1021@mail.ustc.edu.cn}
\orcid{0009-0004-3822-9094}
\affiliation{%
  \institution{University of Science and Technology of China}
  \country{China}
}

\author{Yuqi Zhou}
\email{sasuke18@mail.ustc.edu.cn}
\orcid{0009-0009-5656-8475}
\affiliation{%
  \institution{University of Science and Technology of China}
  \country{China}
}

\author{Bailin Deng}
\orcid{0000-0002-0158-7670}
\email{DengB3@cardiff.ac.uk}
\affiliation{%
  \institution{Cardiff University}
  \country{United Kingdom}}

\author{Juyong Zhang}
\email{juyong@ustc.edu.cn}
\orcid{0000-0002-1805-1426}
\authornote{Corresponding author (\href{mailto:juyong@ustc.edu.cn}{juyong@ustc.edu.cn}).}
\affiliation{%
  \institution{University of Science and Technology of China}
  \country{China}
}

\usepackage[ruled]{algorithm2e} 

\SetAlFnt{\small}
\SetAlCapFnt{\small}
\SetAlCapNameFnt{\small}
\SetAlCapHSkip{0pt}

\usepackage{enumitem}
\usepackage{makecell}
\usepackage{wrapfig}
\usepackage{subcaption}
\usepackage{enumitem}
\usepackage{makecell}
\usepackage{wrapfig}
\usepackage{subcaption}
\usepackage{setspace}
\usepackage{bm}
\usepackage{algorithmic}
\begin{document}

\title{Shape from Semantics: 3D Shape Generation from Multi-View Semantics}



\begin{abstract}
Existing 3D reconstruction methods utilize guidances such as 2D images, 3D point clouds, shape contours and single semantics to recover the 3D surface, which limits the creative exploration of 3D modeling. In this paper, we propose a novel 3D modeling task called ``Shape from Semantics'', which aims to create 3D models whose geometry and appearance are consistent with the given text semantics when viewed from different views. The reconstructed 3D models incorporate more than one semantic elements and are easy for observers to distinguish. We adopt generative models as priors and disentangle the connection between geometry and appearance to solve this challenging problem. Specifically, we propose Local Geometry-Aware Distillation (LGAD), a strategy that employs multi-view normal-depth diffusion priors to complete partial geometries, ensuring realistic shape generation. We also integrate view-adaptive guidance scales to enable smooth semantic transitions across views. For appearance modeling, we adopt physically based rendering to generate high-quality material properties, which are subsequently baked into fabricable meshes. Extensive experimental results demonstrate that our method can generate meshes with well-structured, intricately detailed geometries, coherent textures, and smooth transitions, resulting in visually appealing 3D shape designs.
\end{abstract}



\begin{teaserfigure}
 \includegraphics[width=\textwidth]{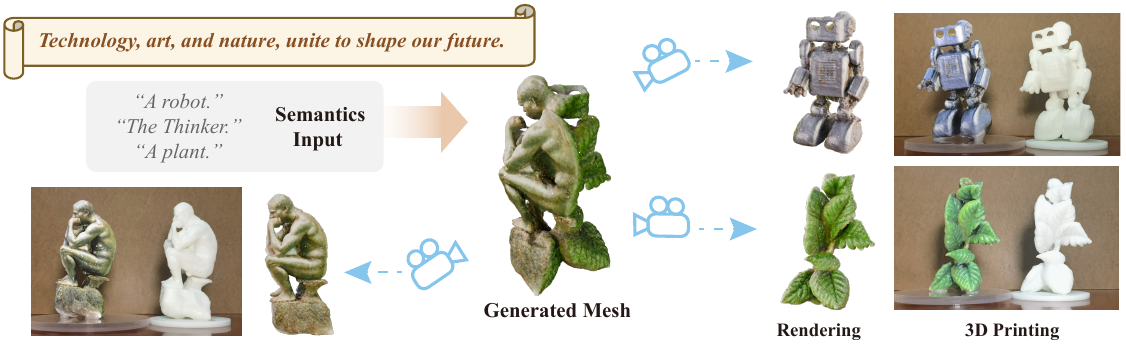}
 \caption{We propose and address \textbf{Shape from Semantics}, a novel generative problem. Given a set of semantics and corresponding views as input, our method can produce high-quality shapes that exhibit geometry and appearance consistent with the semantics from each view and are feasible for real-world fabrication.}
 \label{fig:teaser}
\end{teaserfigure}
\maketitle

\section{Introduction}
\label{intro}

Reconstructing 3D shapes from various inputs is a cornerstone of computer graphics and vision, vital for applications like cultural heritage, film, and architecture. Conventional methods reconstruct 3D geometry and renderings using specific visual data such as RGB images, point clouds, or surface normals. While these inputs facilitate accurate surface reconstruction, they impose strict constraints that can limit the generation of imaginative and novel 3D assets crucial for design, AR/VR, and art.

In this paper, we introduce a novel "Shape from Semantics" problem, which utilizes textual descriptions to guide 3D model generation. Here, semantics are high-level, human-interpretable concepts describing an object's desired characteristics. This semantic-driven approach enables the creation of 3D models that convey intended visual properties from multiple viewpoints, offering a new level of flexibility and creativity in 3D content generation. The resulting 3D models provide a more intuitive and immersive experience through direct observation from different views, compared to 2D designs or projections. Furthermore, semantics-based operations are inherently user-friendly, empowering even non-professionals to produce detailed meshes and intricate textures from just a few text prompts, thereby significantly lowering the barrier to artistic creation.

This task is non-trivial as it requires matching geometry and appearance with input semantics from different viewpoints rather than specific input images; this makes existing multi-view reconstruction techniques inapplicable. Current text-to-3D generation models are also unsuitable, as they typically use a single prompt to describe a single object, whereas our method employs multiple prompts to define different object appearances from multiple views.
The research most similar to ours uses information like shadows or 2D contours as guidance for reconstruction and design. For instance, Shadow Art~\cite{MP09} designs objects whose projections match given 2D shapes under specific lighting. Wire Art~\cite{Hsiao:2018:MVWA:,qu2023dreamwire,Fabricable3DWireArt} focuses on generating wireframe geometries that align with 2D line drawings or outline shapes consistent with semantic inputs. However, these methods primarily offer a two-dimensional visual experience; directly observing the 3D objects often makes it challenging to perceive the intended embedded semantic information. Additionally, such techniques frequently depend on specific setups (e.g., light sources, projection planes) and face fabrication challenges, limiting their practical use.

To address our challenging problem, we leverage the text understanding capabilities of generative models to create a 3D model that matches input semantics from different observation directions. Our approach disentangles the generation process into separate geometry and appearance stages. For geometry, our core insight is that required geometric parts are derived from complete geometries corresponding to the input semantics, which motivates the use of 3D-consistent priors. To this end, we propose Local Geometry-Aware Distillation (LGAD), a strategy employing a multi-view normal-depth diffusion model~\cite{qiu2024richdreamer} as a prior to construct high-quality geometry, represented using Tetrahedron Splatting~\cite{tetsp}. We also introduce a view-adaptive guidance scale to promote smooth semantic transitions across views.
For appearance, we employ a physically based rendering (PBR) pipeline, and utilize a Depth-conditioned Albedo diffusion model to generate and bake high-quality material properties into the fabricable meshes.

Extensive experiments demonstrate our method's capacity for high creativity, generating models that surpass traditional spatial intuition or non-semantic inputs. The resulting 3D models feature well-structured, intricately detailed geometry, coherent textures, and smooth transitions, presenting fascinating and surprising creative designs.
In summary, our contributions include:
\begin{itemize}[leftmargin=*]
    \item 
   We introduce a novel ``Shape from Semantics'' problem for 3D generation from semantics of different views, which provides a powerful modeling tool for design and artistic creation.
    \item 
    We propose Local Geometry-Aware Distillation for robust 3D geometry from limited per-semantic views by directly guiding local normal-depth features with a 3D prior; a view-adaptive guidance strategy for coherent multi-semantic integration; and a PBR-based appearance modeling approach utilizing an albedo diffusion prior to generate high-quality textures.
    \item 
    Our method enables creating high-quality meshes with detailed textures and rich geometry from just a few prompts. 
\end{itemize}

\section{Related work}
\paragraph{Shape from X} Traditional ``Shape from X'' methods focus on high-precision reconstruction of existing objects using known specific visual data, such as RGB images \cite{colmap1,colmap2,SFM,MVS,WLL*21}, depth \cite{kfusion,bfusion} and normals \cite{bini2022cao,Kadambi15}. 
A related body of research explores constructing single, fixed objects that offer multiple visual interpretations; these methods achieve diverse visual perceptions by leveraging factors such as viewing distance \cite{Hybridimages}, figure-ground organization \cite{GeneratingAmbiguousFigure-GroundImages}, illumination from different directions \cite{Reliefsasimages,ShadowPix,Attenuators}, light reflections \cite{reflectors,Mirror}, viewing angles \cite{Generationofviewdependentmodels,LenticularObjects,Hsiao:2018:MVWA:,Fabricable3DWireArt,qu2023dreamwire}, and shadow casting on external planar surfaces \cite{MP09,STR22}. 
However, the goal of these works is typically to produce different 2D information perceptions from an object\textemdash{}whether as a contour\cite{Hsiao:2018:MVWA:,Fabricable3DWireArt,qu2023dreamwire}, a projection \cite{MP09,STR22}, or a picture \cite{Mirror,caustics,softshadow}. Our work enables direct 3D perception, allowing the characteristics of 3D objects to be experienced firsthand. This is the first work to explore creating multiple 3D interpretations of a single object. In addition, we leverage semantics as a substitute for traditional inputs, similar to \cite{Fabricable3DWireArt,qu2023dreamwire}, significantly expanding the creative space. 

\paragraph{3D Data Representations.} The representation of 3D data is a core topic in computer graphics and vision. Beyond traditional point cloud and mesh representations, many novel 3D representations have recently demonstrated significant advantages. \citet{MST*21} propose Neural Radiance Fields (NeRF), which represent a scene with a neural implicit function guided by neural rendering. NeRFs have been widely applied to multi-view reconstruction \cite{WLL*21,neus2,li2023neuralangelo}, sparse reconstruction \cite{Niemeyer2021Regnerf,yu2020pixelnerf,Jain_2021_ICCV,wynn-2023-diffusionerf,liu2023zero1to3}, and generation tasks \cite{jain2021dreamfields,poole2022dreamfusion,wang2023prolificdreamer,lin2023magic3d,chen2023fantasia3d,MAkeit3D}, thanks to its capability in representing objects with rich details. However, its optimization can be time-consuming and computationally intensive. Recently, 3DGS \cite{3DGauss} brings new possibilities for rendering \cite{Yu2023MipSplatting,mssplatting,scaffoldgs} and reconstruction problems \cite{dynamicgs,CFGS,zhu2023FSGS,guedon2023sugar} thanks to its flexible model design and efficient differentiable rendering framework. \citet{tang2023dreamgaussian} incorporate a generative model into 3DGS, enabling rapid generation of textured meshes. However, the geometry generated by 3DGS often suffers from significant detail loss, excessive surface undulations, and suboptimal mesh quality.


Other  representations~\cite{mosdf,guo2024tetsphere} have also demonstrated advantages in reconstruction and generation tasks. DMTET~\cite{shen2021dmtet} combines implicit and explicit representations by predicting surfaces on a deformable tetrahedral grid and extracting meshes via Marching Tetrahedra, enhancing accuracy and efficiency. Fantasia3D~\cite{chen2023fantasia3d} successfully applies this representation to 3D generation tasks. Tetrahedron Splatting~\cite{tetsp} combines precise mesh extraction enabled by tetrahedral grids with efficient optimization of volumetric rendering and demonstrates outstanding performance in generation tasks. In this work, we utilize this geometric representation to achieve high-fidelity geometry and detailed textures while  reducing computational costs.

\paragraph{3D Generation.} While generative models have recently gained widespread attention in computer vision and graphics, the task of 3D generation continues to pose substantial challenges, primarily due to the limited availability of extensive, high-quality 3D datasets~\cite{ABC,wu2023omniobject3d,objaverse,objaverseXL}. Recently, many 3D generation methods \cite{jain2021dreamfields,poole2022dreamfusion,wang2023prolificdreamer,lin2023magic3d,chen2023fantasia3d,MAkeit3D} utilize 2D information as supervision to guide 3D generation, using various representations of 3D data. DreamFields \cite{jain2021dreamfields} pioneers the use of diffusion models for semantic-based 3D generation. DreamFussion~\cite{poole2022dreamfusion} introduces the score distillation sampling (SDS) loss, which leverages semantic information and 2D rendering results, and this approach has since been widely adopted. However, as these methods inherently rely on supervision from 2D rendering results, they often face challenges with multi-view inconsistency. While many existing works aim to mitigate such inconsistencies \cite{liu2023zero1to3,shi2023MVDream}, our approach leverages such potential inconsistency to generate creative objects with multiple visual interpretations. Moreover, researchers incorporate various priors (normal, depth, etc.) into 3D generation tasks to enhance the realism of models. SweetDreamer~\cite{sweetdreamer} and RichDreamer \cite{qiu2024richdreamer}~integrate canonical coordinate maps and normal-depth priors into the loss function, respectively. Meanwhile, Wonder3D \cite{long2023wonder3d} and CRM \cite{wang2024crm} directly utilize these priors to construct corresponding meshes. 

Researchers also try to use 3D datasets directly for 3D generation tasks. PolyGen~\cite{polygen}, MeshGPT~\cite{siddiqui_meshgpt_2024}, and XCube~\cite{ren2024xcube} represent geometry natively using mesh vertices, mesh surface sequences, and voxels, respectively. SDFusion~\cite{cheng2023sdfusion} and 3DGen~\cite{3Dgen} leverage 3D Variational Autoencoders (VAEs) to encode geometry, employing Signed Distance Fields (SDFs) and triplanes as geometric representations. Methods such as Shap-E~\cite{shapee} and 3DShape2VecSet~\cite{3dshape2vec} adopt transformer-based architectures to encode geometry, while more recent methods such as TRELLIS~\cite{trellis} and CLAY~\cite{zhang2024clay} focus on constructing more compact and versatile latent spaces for decoding into diverse representations, DeepMesh~\cite{zhao2025deepmesh} and OctGPT~\cite{wei2025octgpt} enhance pretraining efficiency and stability via autoregressive modeling.
 However, they are typically trained and evaluated on datasets such as ShapeNet~\cite{ShapeNet} and Objaverse~\cite{objaverse}, which constrains the diversity and complexity of the generated shapes. In contrast, our method seeks to unlock the creative potential of generative models to synthesize astonishing geometric forms that transcend common objects.

\section{Method}
We take as input  $n$ semantic labels $\mathcal{Y} = \{y_i\}$, each being a textual prompt, and their corresponding view directions $\mathcal{V} = \{v_i\in \mathrm{SO\left(3\right)}\}$. We call  $\mathcal{V}$ the \emph{observation views}, which can either be predefined or initialized randomly. We aim to generate a colored 3D shape $\mathcal{S}$ whose texture and geometry align with the associated semantic class $C\left(y_{i}\right)$ when observed from any main review $v_{i}$. 
$\mathcal{S}$ should possess a simple, intuitive, and compact design suitable while retaining key geometric features that define its appearance. Meanwhile, the generated shape should be highly recognizable and visually elegant.

{\newcommand{\cropablationA}[1]{
  \makecell{
  \includegraphics[height=0.2\linewidth]{#1} 
  }
}

\begin{figure}[t!]
\centering
\setlength{\tabcolsep}{1pt}
\renewcommand\arraystretch{0.7}
\resizebox{\linewidth}{!}{
\begin{tabular}{c|ccc}
    \toprule
    \cropablationA{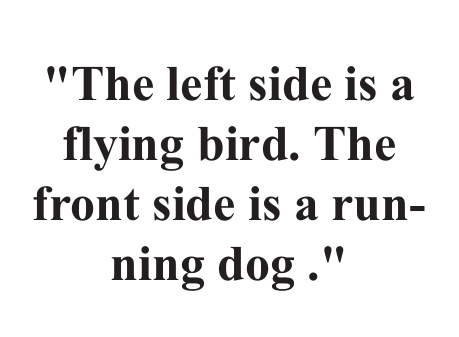} &
    \cropablationA{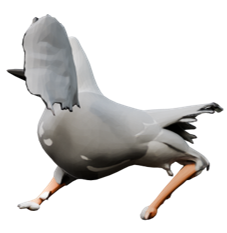} &
    \cropablationA{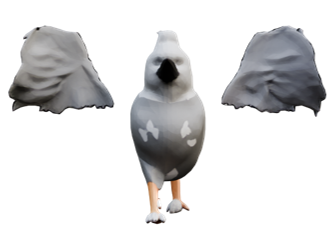} &
    \cropablationA{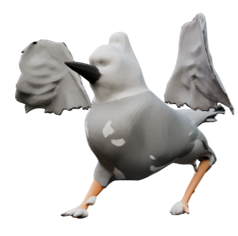} \\
    \midrule
    \cropablationA{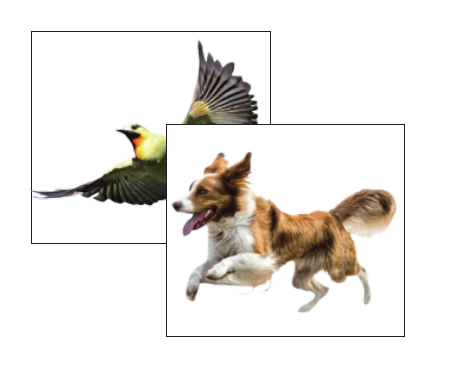} &
    \cropablationA{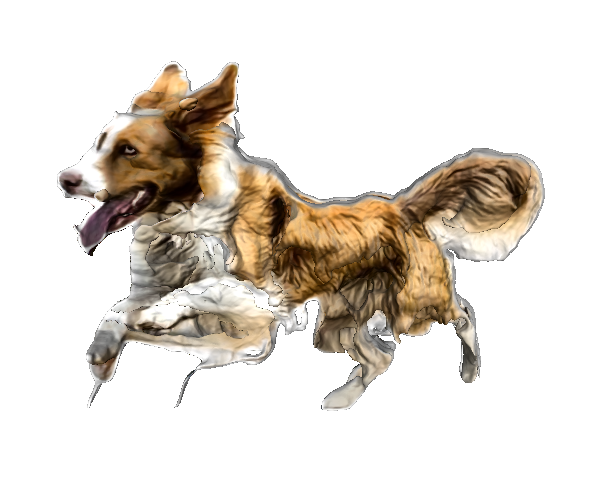} &
    \cropablationA{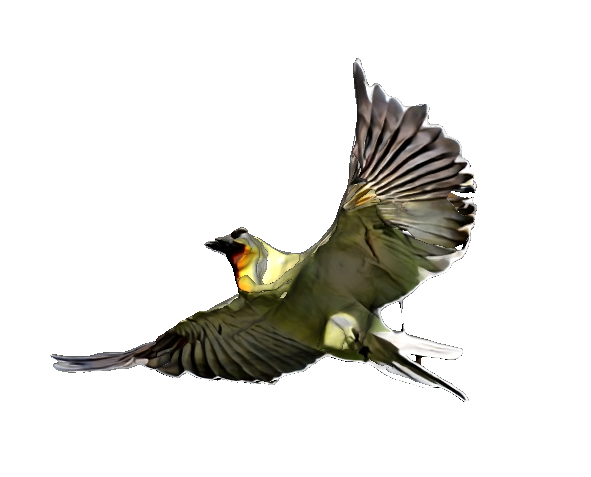} &
    \cropablationA{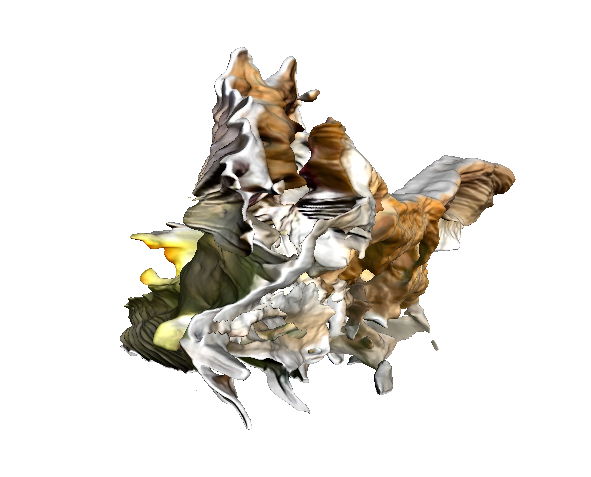} \\
    \midrule
    Input & Front & Left& Random View\\
    \bottomrule
\end{tabular}
}
\caption{\textbf{Failure of Naive Solutions.} We expect the shape to combine ``a running dog'' (front side) and ``a flying bird'' (left side). The top row shows the text-to-3D generation result from TRELLIS~\cite{trellis}, which mixes the semantics directly. The bottom row shows a baseline approach that first generates 2D images using Stable Diffusion~\cite{sd35}, then performs multi-view reconstruction~\cite{WLL*21}, whose result shows meaningless geometry. Our result of this case can be found in Fig.~\ref{fig:wireArt}.}
\label{fig:naive_failure}
\end{figure}
}
\begin{figure*}
    \centering
    \includegraphics[width=\textwidth]{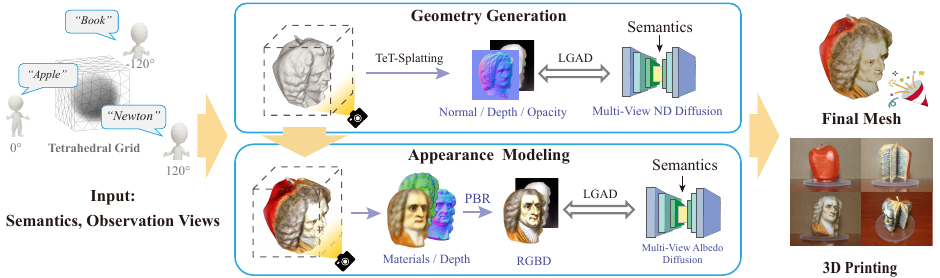}
    \caption{\textbf{Shape from Semantics Pipeline.} We use TeT-Splatting~\cite{tetsp} as the 3D representation, and disentangle geometry and appearance generation into a two-stage process. In the geometry generation stage, we render the normal and depth map through alpha blending, and optimize the geometry using the proposed LGAD method. In the appearance modeling stage, we use physically based rendering to obtain the RGBD map for diffusion and learn view-independent realistic textures. Finally, the colored mesh is extracted and can be crafted into visually appealing art pieces.}
    \label{fig:pipeline}
\end{figure*}

This task is inherently challenging. Despite recent advances in generative models, their direct application to our problem yields unsatisfactory results. 
For instance, state-of-the-art text-to-3D models struggle with our task (Fig.~\ref{fig:naive_failure}, top) due to their limited understanding of directional descriptions, leading to semantic blending across different views.
An alternative approach involves generating an image for each semantic label using a text-to-image model (e.g., Stable Diffusion~\cite{sd35}) followed by 3D reconstruction from these multi-view images~\cite{WLL*21}.
However, this often results in meaningless or distorted geometries (Fig.~\ref{fig:naive_failure}, bottom) because the generated images lack the necessary 3D geometric information for robust reconstruction, leading to flattened or deformed shapes.

To overcome these limitations, we propose a novel solution (see Fig.~\ref{fig:pipeline}) that disentangles geometry and appearance generation into a two-stage process, ensuring geometrically plausible and semantically coherent results. Section~\ref{3_1} introduces our Local Geometry-Aware Distillation (LGAD) approach that leverages geometric priors from pre-trained diffusion models to achieve high-quality geometry under the limited view range for each semantic constraint. Then, Section~\ref{3_2} presents our 3D geometry representation and complementary strategies for effective geometry generation. Finally, Section~\ref{3_3} describes our PBR approach for appearance modeling, which produces high-quality, fabricable textures.

\subsection{Local Geometry-Aware Distillation}
\label{3_1}
\begin{figure}[t]
  \centering
  \includegraphics[width=\linewidth]{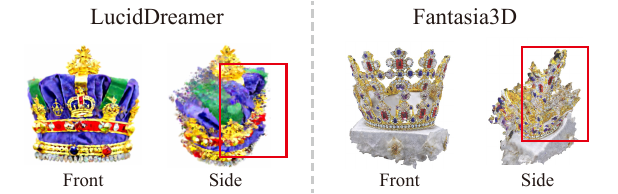}
  \caption{\textbf{SDS under Limited Sampling Views}. To simulate the constraints we encounter, we reduced the SDS sampling azimuth range from $[-180^{\circ}, 180^{\circ}]$ to $[-45^{\circ}, 45^{\circ}]$. The semantics is ``an imperial state crown of England''. While LucidDreamer~\cite{liang2024luciddreamer} with modified SDS achieves high-quality frontal rendering, significant shape collapse and fragmentation become evident upon rotation (marked by red box). Even with normal maps applied for geometric optimization~\cite{chen2023fantasia3d}, the shape still suffers from structural collapse/deformation.}
  \label{fig:limited_views}
\end{figure}

Given the absence of suitable datasets for our novel task, training a data-driven 3D generative model directly is infeasible. Therefore, we adopt the score distillation sampling (SDS) framework~\cite{poole2022dreamfusion} to leverage powerful pre-trained diffusion models as priors for 3D shape generation.
In a typical SDS iteration, a camera pose $c$ and corresponding semantics $y(c)$ are sampled. An image $\mathcal{I}=\mathcal{I}(\theta, c)$ is rendered from the current 3D shape representation $\theta$. This image is then guided by a text-to-image diffusion model to match $y(c)$. The SDS gradient is commonly expressed as:
\begin{equation}
    \nabla_{\theta} \mathcal{L}_{\mathrm{SDS}}(\theta) \triangleq\mathbb{E}_{t, \boldsymbol{\epsilon}, c}\left[\omega(t)\left(\boldsymbol{\epsilon}_{pre}\left(\mathcal{I}_t; t, y(c)\right)-\boldsymbol{\epsilon}\right) \frac{\partial \mathcal{I}}{\partial \theta}\right],
    \label{eq:sds}
\end{equation}
where $\mathcal{I}_t$ is the noised rendered image with sampled noise $\boldsymbol{\epsilon}$, $\boldsymbol{\epsilon}_{pre}$ is the noise predicted by the 2D diffusion prior, and $\omega(t)$ is a  weighting function.
However, effective SDS relies on dense and varied view sampling. Our problem inherently restricts the view range for each semantic, as multiple distinct semantics must be expressed from specific viewpoints of a single object. This leads to weak geometric supervision, often resulting in a mismatch between the intended shape and the rendered appearance, or incorrect details (Fig.~\ref{fig:limited_views}).

To address this challenge and achieve robust geometry under limited-view supervision, we introduce \emph{Local Geometry-Aware Distillation} (LGAD). The core principle is that any plausible local geometric attribute (e.g., surface normals and depth observed from a viewpoint $v_i$) corresponding to a semantic $y_i$ must be consistent with some complete 3D shape $\theta^*$ that fully embodies $y_i$. For a given semantic $y(c)$ from a camera view $c$, let $\boldsymbol{g} = P(\theta,c)$ be the currently rendered local geometric attributes (specifically, normal and depth maps, or ND maps) from our evolving shape $\theta$. LGAD aims to guide $\theta$ such that $g$ aligns with the local attributes $\boldsymbol{g}^*=  P(\theta^*,c)$ that would be observed from such an ideal shape $\theta^*$ along view $c$.

Rather than explicitly reconstructing $\theta^*$, LGAD uses a pre-trained 3D-aware diffusion model as a prior. The key is to shift the distillation target from 2D RGB images to ND maps. As our 3D-aware prior, we employ the multi-view normal-depth diffusion model from RichDreamer~\cite{qiu2024richdreamer}, which is represented as a noise prediction network $\boldsymbol{\epsilon}_{pre}^{3D}$ conditioned on the semantic $y(c)$ and a set of camera views $C$ (which includes the observation view $c$ and other views surrounding the object). We then formulate a loss that measures the deviation between the predicted and ground-truth noises for the view $c$, similar to the SDS loss in Eq.~\eqref{eq:sds}.
To satisfy the RichDreamer prior's requirement for multi-view ND inputs for all views in $C$ while focusing on guidance from $y(c)$, we render a single noise-free ND map $\boldsymbol{g}_0(\theta, c)$ from the observation view $c$, and duplicate it for each view in $C$ with individually added noise per view, creating a set of noised maps $\{\boldsymbol{g}_{t,c'}\}_{c' \in C}$ that shared the same underlying geometry. This set is then input to $\boldsymbol{\epsilon}_{pre}^{3D}$ along with the semantics $y(c)$ and the views $C$. We then use the noise prediction $\left.\boldsymbol{\epsilon}_{pre}^{3D}\left(\{\boldsymbol{g}_{t,c'}\}_{c' \in C}; t, y(c), C\right)\right|_c$ 
corresponding to the observation view $c$ and compare it with the ground truth $\boldsymbol{\epsilon}_c$. Our final LGAD loss gradient is:
\begin{equation}
    \nabla_{\theta} \mathcal{L}_{} =\mathbb{E}_{t, \boldsymbol{\epsilon}, c}\left[\omega(t)\left(\left.\boldsymbol{\epsilon}_{pre}^{3D}\left(\{\boldsymbol{g}_{t,c'}\}_{c' \in C}; t, y(c), C\right)\right|_c-\boldsymbol{\epsilon}_{c}\right) \frac{\partial \boldsymbol{g}}{\partial \theta}\right].
    \label{eq:lgad_2}
\end{equation}   
A detailed pseudo-code is shown in the supplementary materials.

\begin{figure}[t]
  \centering
  \includegraphics[width=\linewidth]{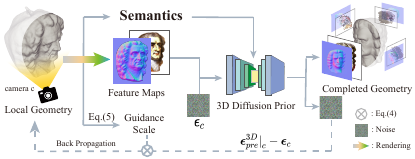}
  \caption{\textbf{Local Geometry-Aware Distillation.} In each iteration, we sample a camera and render feature maps of the local geometry. Semantics and guidance scale are obtained through Eq.~\eqref{eq:scale}. Afterwards, 3D diffusion prior is utilized to denoise the features to match the front view feature of a complete geometry, and finally back propagate to optimize the local geometry.}
  \label{fig:lgad}
\end{figure}

\subsection{Geometry Representation and Generation}
\label{3_2}
\paragraph{Tetrahedron Splatting.} In 3D generation tasks, implicit representations like NeRF~\cite{MST*21} can involve lengthy training, while explicit representations such as 3DGS~\cite{3DGauss, tang2023dreamgaussian} may produce unstructured or low-quality geometry. Instead, we follow~\cite{tetsp} and adopt tetrahedral splatting as our representation, which constructs a tetrahedral grid encoding a Signed Distance Field (SDF) in 3D and uses alpha blending for tetrahedron rendering. To enhance geometric quality during training, we incorporate an eikonal loss and a normal consistency loss:
\begin{equation}
    \mathcal{L}_{\text {eik}}=\sum\nolimits_{\delta}\left(\left\|\nabla f_\delta\right\|_{2}-1\right)^{2}, \mathcal{L}_{\mathrm{nc}}=\sum\nolimits_{e}\left(1-\cos \left(\boldsymbol{n}_{e_{1}}, \boldsymbol{n}_{e_{2}}\right)\right),
\end{equation}
where $\nabla{f}_{\delta}$ is the SDF gradient of each tetrahedron $\delta$, and $\boldsymbol{n}_{e_{1}}$ and $\boldsymbol{n}_{e_{2}}$ are the surface normals at the vertices connected by edge grid $e$.

\begin{figure}[t]
  \centering
  \includegraphics[width=\linewidth]{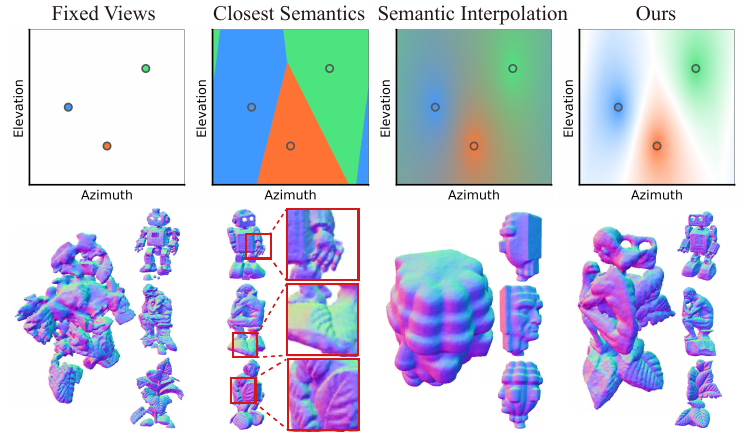}
  \caption{\textbf{View-Adaptive Guidance}. The top row shows the guidance scale variation across camera views, with circles marking observation views. Point colors denote semantic components, while transparency indicates the guidance scale. Four strategies are tested on the case in teaser: training on fixed observation views and their semantics, with randomly sampled views and their closest semantics, semantics interpolation, and our method. The bottom row shows normal maps of generated geometries, with the second case contains three middle views, and the other contain three observation views and a random view. Training with fixed views leads to fractured structures, semantics interpolation makes shape less expressive, and directly choosing closet semantics causes geometric feature blending at semantic boundaries, exemplified by generating a human hand for the robot, or plant leaves on the sculpture and the robot body.}
  \label{fig:guide}
\end{figure}

\paragraph{View-Adaptive Guidance.} 
While our LGAD loss induces strong geometric supervision, simply applying it at the predefined observation views $v_i \in \mathcal{V}$ often fails to produce satisfactory overall 3D geometry (see Fig.~\ref{fig:guide}, left). For a coherent shape, it is beneficial to also apply LGAD guidance from camera views that are near the observation views. However, this introduces a challenge: the input semantics are explicitly defined only for the observation views; for any other views $c \notin \mathcal{V}$, the intended semantics $y(c)$ are ambiguous.

A naive approach is to assign $y(c)$ based on the semantics of the closest observation view. However, this leads to abrupt and potentially conflicting transitions in areas where the influence of two observation views meets (Fig.~\ref{fig:guide}, second column). An alternative is to interpolate the embeddings of the surrounding observation view semantics $\{y_i\}$ by weighting them based on proximity to $c$, e.g., $\text{Emb}[y(c)] = \left(\sum\nolimits_i w_i\cdot \text{Emb}[y_i]\right)/\left(\sum\nolimits_i w_i\right)$, where $w_i = 1/(1-c\cdot v_i)$ are influence weights, and $\text{Emb}(\cdot)$ is the text encoding function. As shown in Fig.~\ref{fig:guide} (third column), this semantic blending significantly decreases the geometric expressiveness and distinctiveness of the intended multiple interpretations. 
To address these issues, we propose a View-Adaptive Guidance strategy that utilizes Classifier-Free Guidance (CFG)~\cite{cfg}. CFG allows modulation of the semantic guidance strength via a scale parameter $s$:
\begin{equation}
\tilde{\boldsymbol{\epsilon}}_{pre}\left(\mathcal{I}_{t}, t, y\right):=s\boldsymbol{\epsilon}_{pre}\left(\mathcal{I}_{t}, t, y\right)+(1-s)\boldsymbol{\epsilon}_{pre}\left(\mathcal{I}_{t}, t, \varnothing\right),
  \label{eq:cfg}
\end{equation}
where $\tilde{\boldsymbol{\epsilon}}_{pre}$ is the guided noise prediction, ${\boldsymbol{\epsilon}}_{pre}$ is the model's raw noise prediction (conditioned on semantics $y$ or an unconditional prompt $\varnothing$), and $\mathcal{I}_{t}$ is the noised input. We can dynamically adjust  $s$ to enforce stronger guidance when the camera view $c$ is closer to an observation view, and weaker guidance when it is in an ambiguous transition zone. Specifically, we sort the influence weights of each observation view on the current view $c$ in descending order: $\{w^{\prime}_{0} \geq \ldots \geq w^{\prime}_{n-1}\}$, and compute the guidance scale as:
\begin{equation}
    s = s_0(w^{\prime}_0-w^{\prime}_1)/\sum\nolimits_i w^{\prime}_i,
    \label{eq:scale}
\end{equation}
where $s_0$ is a hyperparameter. This ensures $s$ is largest when $c$ aligns with a single observation view and diminishes as $c$ moves into regions where multiple observation views have comparable influence. Additionally, to avoid semantic blending in the prior conditioning, we provide the LGAD diffusion prior with the semantics corresponding to the observation view closest to $c$. Fig.~\ref{fig:guide} shows that when the dominant semantic influence transitions from one observation view to another, $s$ naturally passes through or near zero. This creates continuous and smooth semantic supervision across views, leading to more coherent and expressive geometric results.

\paragraph{Training Details.} Our geometry generation employs a structure-to-detail process. We initialize the tetrahedral SDF field as a sphere, then apply LGAD to obtain a coarse geometry. Afterwards, 
for detailed geometric refinement, we lower the timestep sampling range in the diffusion process. 
Throughout the training process, we also integrate vanilla Stable Diffusion~\cite{LatentDiffusion} as an additional guidance complementing our LGAD optimization: 
\begin{equation}
    \nabla_{\theta} \mathcal{L}_{\mathrm{SDS}}=\mathbb{E}_{t, \boldsymbol{\epsilon}, c}\left[\omega(t)\left(\boldsymbol{\epsilon}_{\phi}(\mathcal{I} ; y(c), t)-\boldsymbol{\epsilon}\right) \frac{\partial \mathcal{I}}{\partial \theta}\right],
    \label{eq:true_sds}
\end{equation}
where $\mathcal{I}_t$ is the noised rendered normal maps from view $c$, and $\boldsymbol{\epsilon}_{\phi}(\mathcal{I} ; y(c), t)$ is the noise estimated by the UNet $\epsilon_{\phi}$ of the 2D prior.

\subsection{Appearance Modeling}
\label{3_3}
With the geometry established, this stage focuses on adding rich color and realistic surface appearance to the 3D model. To this end, the well-trained tetrahedral SDF field is first converted into a polygonal mesh using the Marching Tetrahedra algorithm~\cite{shen2021dmtet}. For the subsequent appearance optimization, we employ Physically Based Rendering (PBR) better disentangle material properties and achieve more realistic results.
The material properties at any surface point $p$ are determined by the diffuse color $k_{d} \in \mathbb{R}^{3}$ (albedo), roughness $k_{r} \in \mathbb{R}$, metallic term $k_{m} \in \mathbb{R}$, and tangent-space normal variation $k_{n} \in \mathbb{R}^{3}$. These spatially varying attributes are encoded using a hash grid $\Phi_\Theta$ with parameter $\Theta$:
$\left(k_{d}, k_{r}, k_{m}, k_{n}\right)=\Phi_{\Theta}(p)$.
To ensure the final baked appearance on the extracted mesh is high-fidelity and consistent with training renders, our PBR setup decouples materials from view-dependent lighting effects, making all physical attributes spatially invariant for faithful extraction.

We use a Depth-conditioned Albedo diffusion model~\cite{qiu2024richdreamer} as the appearance prior, capable of producing multi-view albedo maps conditioned on semantics and camera poses. The LGAD framework from Sec.~\ref{3_1} is adapted to optimize these PBR materials by using rendered RGBD images as the distillation target $\boldsymbol{g}$ in Eq.~\eqref{eq:lgad_2}. Additionally, the view-adaptive guidance strategy and auxiliary SDS loss from Sec.~\ref{3_2} are utilized to further refine the appearance.




\section{Experiments}
\paragraph{Implementation Details}
During training, camera views are sampled with an azimuthal range of $\pm50$ degrees around each observation view and an elevation range of $\pm25$ degrees, further enhanced with adaptive scale adjustments in Eq.~\eqref{eq:scale} with $s_0 = 70$. The geometry generation stage takes 3,000 iterations, which includes 1,000 iterations for initial coarse shape formation and 2,000 iterations for subsequent geometric refinement. Appearance modeling is then performed for an additional 2,000 iterations. The entire training procedure is performed on a single NVIDIA RTX 3090 GPU (24GB VRAM) and completes in approximately 1.5 hours. We maintain a consistent tetrahedral grid resolution of $256^3$ for both the training and mesh extraction stages.

\begin{figure}[t]
  \centering
  \includegraphics[width=\linewidth]{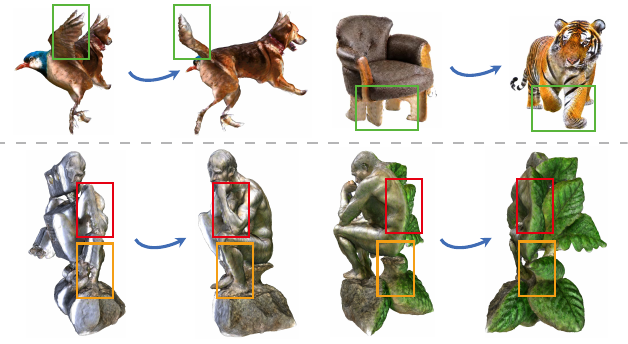}
  \caption{\textbf{Geometric Structures Details.} Our generated results achieve sophisticated visual effects by sharing geometric elements across different semantic components. }
  \label{fig:idea}
\end{figure}

\begin{figure*}[t]
  \centering
  \includegraphics[width=0.9\linewidth]{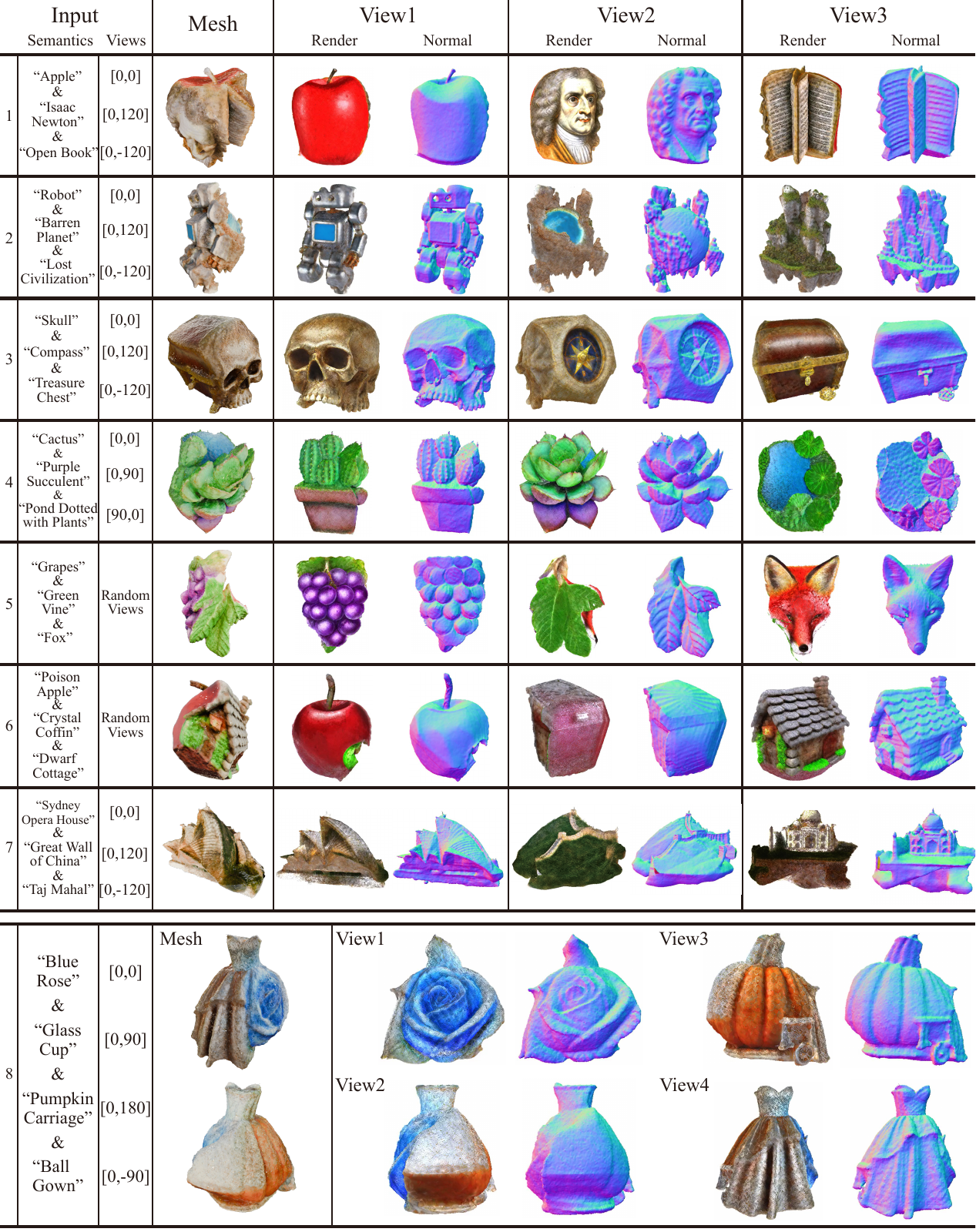}
  \caption{\textbf{Gallery of Shape from Semantics}. We show the inputs, the generated colored mesh, the rendering and normal maps of each semantics. The textured meshes are rendered with Blender. The normal maps show that our generated shapes have meaningful geometries aligned with the renderings. Our method can complete generation with inputs of up to four semantics.}
  \label{fig:main}
\end{figure*}

\paragraph{Main Results}
We apply our method to generate various multi-semantic texture meshes, which are shown in Fig.~\ref{fig:main}. Our inputs cover diverse semantic and view inputs, demonstrating the rich creativity of our method. The results are highly consistent with the expected semantics in terms of geometry and appearance. We 3D-print some cases with results provided in the supplementary materials. The fabricated objects are highly consistent with the expected design and exhibit an aesthetic appeal.

A notable feature of our generated results is their sophisticated geometric structures. As demonstrated in Fig.~\ref{fig:idea}, the LGAD technique enables the same geometric components to serve distinct semantic roles. For instance, a bird's wings transform into a dog's tail when rotated, while the stone bench simultaneously serves as a plant leaf. This kind of combination achieves geometric-semantic transitions during rotational observing, exhibiting rich playfulness.


\begin{table}[t]
    \caption{\textbf{CLIP Similarity (\%) (higher is better) between the Semantics and the Observed Meshes in Fig.~\ref{fig:main}.} Each result is displayed as scores of with/without textures.}
    \centering
    \renewcommand\arraystretch{0.7}
    \setlength{\tabcolsep}{2pt}
    \resizebox{\linewidth}{!}{%
        \begin{tabular}{c c c c c c c c c}
        \toprule 
        \textbf{View/Metric} & \textbf{Case 1} & \textbf{Case 2} & \textbf{Case 3} & \textbf{Case 4} & \textbf{Case 5} & \textbf{Case 6} & \textbf{Case 7} \\ \midrule
        \textbf{View1} & 38.84/37.36 & 37.44/35.37 & 35.98/36.74 & 40.63/37.65 & 38.39/34.17 & 36.41/36.36 & 34.68/36.94 \\
        \textbf{View2} & 41.07/37.32 & 30.77/23.95 & 33.36/30.77 & 37.25/36.47 & 31.51/23.23 & 25.29/25.45 & 34.81/32.57 \\
        \textbf{View3} & 36.24/31.81 & 22.27/22.32 & 40.76/39.95 & 30.39/22.11 & 31.74/33.32 & 41.57/37.37 & 39.45/38.31 \\
        \textbf{Mean Score} & 38.72/35.49 & 30.16/27.21 & 36.70/35.82 & 36.09/32.08 & 33.88/30.24 & 34.43/33.06 & 36.31/35.94 \\
        \bottomrule
    \end{tabular}%
    }
    \vspace{-0.2cm}
    \label{tab:clip_score}
\end{table}

\begin{table}[t]
\caption{\textbf{Scores of the User Study for Results in Fig.~\ref{fig:main}.} The rating range is 0-10, with higher scores indicating better results.}
\centering
\renewcommand\arraystretch{0.7}
\resizebox{0.95\linewidth}{!}{%
    \begin{tabular}{c c c c c c c c c}
        \toprule
        \textbf{Metric} & \textbf{Case 1} & \textbf{Case 2} & \textbf{Case 3} & \textbf{Case 4} & \textbf{Case 5} & \textbf{Case 6} & \textbf{Case 7} \\
        \midrule
        w/o Texture & 6.51 & 5.46 & 7.39 & 6.99 & 5.87 & 6.50 & 7.40 \\
        w/ Texture  & 9.35 & 7.47 & 9.38 & 9.47 & 8.94 & 8.54 & 9.44 \\
        Semantic Pref. & 8.82 & 8.96 & 8.80 & 9.04 & 8.82 & 8.82 & 8.92 \\
        Overall Pref.  & 8.61 & 8.87 & 8.71 & 8.87 & 8.96 & 9.00 & 8.83 \\
        \bottomrule
    \end{tabular}
}
\label{tab:user}
\vspace{-0.3cm}
\end{table}

\paragraph{Quantitative Evaluation}
To evaluate the consistency between the generated results and the input semantics, we render the generated 3D models from each observation view and use the CLIP score~\cite{clip} to measure their semantic similarity to the input. Tab.~\ref{tab:clip_score} presents the CLIP scores for textured and non-textured cases. For each observation view, we allow random variations within a 20-degree latitude and longitude range to render the images. The CLIP model then evaluates the captured results 1,000 times, and the average score is taken as the score for that observation view. The results indicate that the generated models effectively convey semantic information, regardless of whether textures are applied. Observers can also discern the semantic representation of these geometric shapes even with slight changes in perspective.

In addition, we conducted a user study to further validate our method. The participants were shown the rendering of our generated models one at a time, and asked to sequentially answer the following questions with a score from 0 to 10:
\begin{itemize}[leftmargin=*]
\item Q1: How well does the textureless rendering match the semantics?
\item Q2: How well does the textured rendering match the semantics?
\end{itemize}
The last two questions focus on participants' preferences between the results of ours and \cite{Fabricable3DWireArt} under the same semantics. A score closer to 10 indicates a stronger preference for our results:
\begin{itemize}[leftmargin=*]
\item Q3: Which result better aligns with semantics? 
\item Q4: Which overall result do you prefer? 
\end{itemize}
We randomly and fairly selected participants, collecting 83 samples. The average scores from each observation view of each case are presented in Tab.~\ref{tab:user}. The results indicate that our 3D model design receives considerable recognition. 


In preference scoring, our overall performance in semantics and aesthetics is significantly higher compared to the method we benchmarked against. This suggests that the combination of our rendering and geometry maintains strong semantic expressiveness. Moreover, when compared to monochrome or outline results, our overall design proved to be highly appealing.

\paragraph{Qualitative Comparison}
As far as we know, no previous research has been conducted with the same purpose as ours. Therefore, we make comparisons with Shadow Art~\cite{MP09} and Wire Art~\cite{Fabricable3DWireArt}. Similar to us, they aim to represent diverse semantics from different views. Initially, we provide our final rendered images at observation views to both of them and compare their results with ours. As shown in Fig.~\ref{fig:ShadowArt}, both methods convey information solely through contours and silhouettes, lacking color representation and meaningful geometric structure. In contrast, our shape representation integrates rendering, enabling the depiction of more intricate and complex shapes. 

We also compare multi-semantic generation quality in Fig.~\ref{fig:wireArt}. Given the same inputs, our generated shapes are more refined while maintaining better geometric-semantic consistency in local details. Additionally, as shown in the first row ``chair''\&``tiger'' case, our design could convey distinct front/back semantics, which is challenging for contour-based representations.

\begin{figure}[t]
  \centering
  \includegraphics[width=\linewidth]{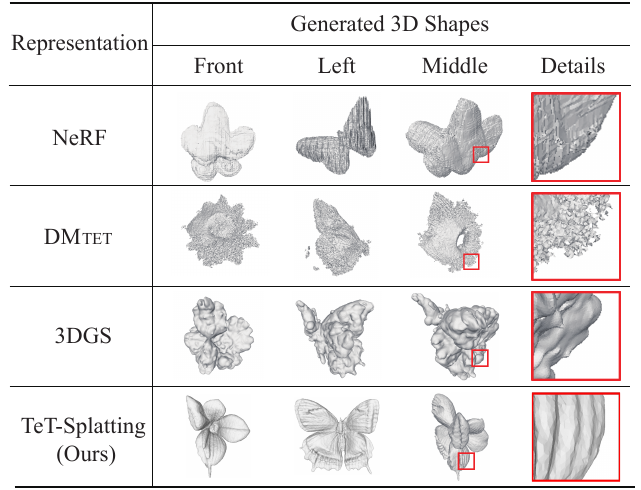}
  \caption{\textbf{Comparison of Different Representations.} The semantics are ``flower''\&``butterfly''. We compare with Dreamfusion~\cite{stable-dreamfusion} using NeRF, Fantasia3D~\cite{chen2023fantasia3d} using \textsc{DMTet}~\cite{shen2021dmtet}, and DreamGaussian~\cite{tang2023dreamgaussian} using 3DGS. We implement the baselines by modifying SDS into multi-semantics version according to Eq~\eqref{eq:sds}. Results show that the TeT-Splatting geometry is smoother and more detailed.}
  \label{fig:presentation}
\end{figure}

Comparisons between different representations are presented in Fig.~\ref{fig:presentation}. The tetrahedral splatting presents superior geometric fidelity, significantly enhancing stability even compared to methods using similar structures like DMTet~\cite{shen2021dmtet, chen2023fantasia3d}.

\begin{figure}[t]
  \centering
  \includegraphics[width=0.95\linewidth]{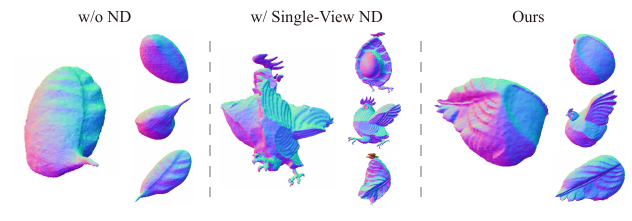}
  \caption{\textbf{Ablation for LGAD.} Three methods use the same randomly sampled observation views, and semantics are [``Fragile Egg'', ``Soaring Chicken'', ``Fallen Feather'']. Here we present normal maps of three observation views and a random view.}
  \label{fig:nd_abl}
\end{figure}

\paragraph{Ablation Study}
To demonstrate the effectiveness of our LGAD strategy, we conduct an ablation study on different score distillation approaches, with results presented in Fig.~\ref{fig:nd_abl}. Results show that the absence of geometric supervision leads to degraded geometry. Using a single-view text-to-ND model, however, results in overfitting, distorted shapes, and the emergence of features from other semantics, such as the chicken claw extending out of the egg. In contrast, our method generates geometry that is rich in features while remaining clean, and faithfully conveying the semantics.

\begin{figure}[t]
  \centering
  \includegraphics[width=0.95\linewidth]{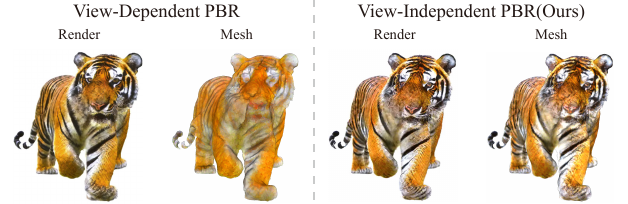}
  \caption{\textbf{Ablation for View-independent PBR.} We run appearance modeling based on the full PBR pipeline and our view-independent PBR on the same generated shape. The former mesh has the albedo extracted as texture.}
  \label{fig:pbr_abl}
\end{figure}
We also conduct an ablation study on different rendering methods, with results presented in Fig.~\ref{fig:pbr_abl}.
Our rendering achieves comparable texture details to the full PBR pipeline. However, as the mesh columns show, our extracted mesh faithfully preserves the rendered appearance, whereas the mesh generated by the full PBR approach (using albedo texture as the extraction basis) exhibits significant detail loss,

In Fig.~\ref{fig:abl_view} we explore how view distributions affect generation. While maintaining the same semantic inputs, we employ two distinct observation view sets for generation. Both groups achieve high-quality rendering and geometry, demonstrating our method's robustness to view variations. Simultaneously, all three semantic instances exhibit distinct shapes, confirming the diversity of our generation approach. The geometry of each semantics adaptively composes spatially coherent structures, achieving geometric compatibility while showcasing rich creativity. 
\section{Conclusion \& Discussion}

We introduced and addressed "Shape from Semantics," a novel problem focused on generating 3D shapes from multi-view semantics. Our core approach leverages 3D diffusion priors for both shape and appearance optimization. Experiments show our method successfully produces impressive shapes that are aesthetically pleasing, semantically consistent with inputs, and readily manufacturable.

\begin{figure*}[t] 
  \begin{minipage}[!t]{0.49\textwidth}
    \centering
    \includegraphics[width=\linewidth]{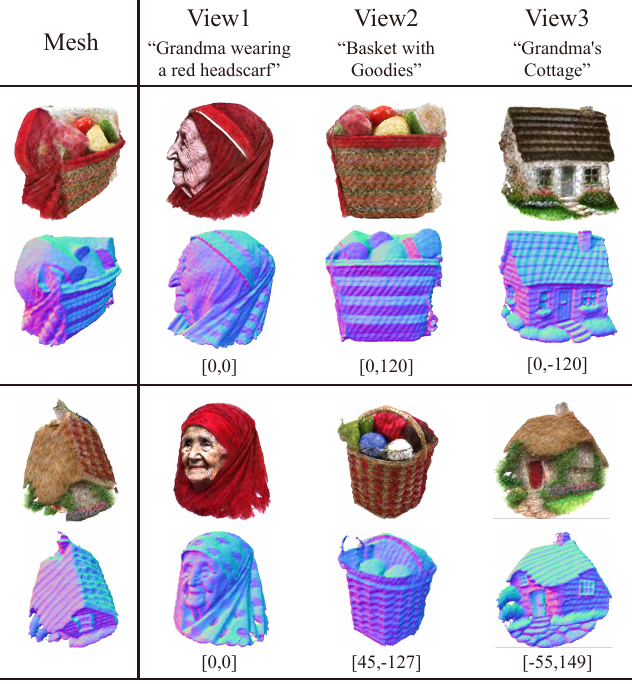}
      \caption{\textbf{Ablation Study on Different Observation Views.} Under fixed semantics, we present RGB and normal maps under an orbiting pattern (top row) and randomly sampled (bottom row). Our method adaptively generates matched shape compositions conditioned on view variations. It can be observed that with orbiting viewpoints, the relative independence of perspectives leads to slightly larger volumetric shapes for each semantic component.}
  \label{fig:abl_view}
  \end{minipage}
  \hfill
  \begin{minipage}[!t]{0.49\textwidth}
    \centering
    \includegraphics[width=\textwidth]{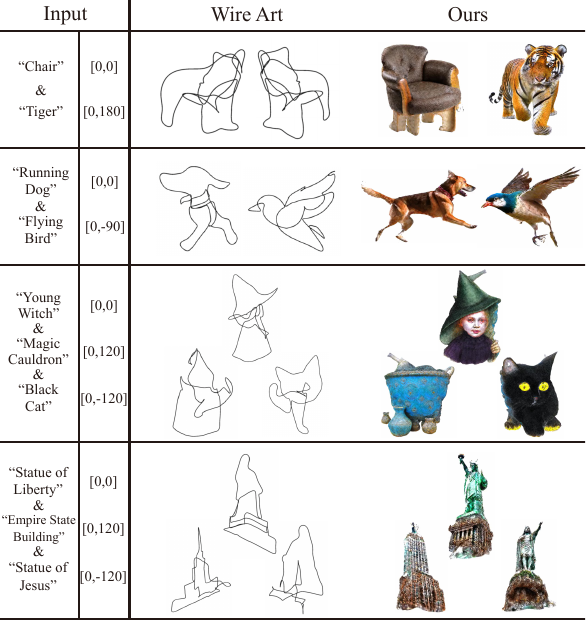}
    \caption{\textbf{Comparison with Wire Art \cite{Fabricable3DWireArt}.} We use the same semantics for comparison. The top-row result highlights the limitations of Wire Art, which arise from its dependence on projections to convey information, therefore, it is difficult to complement back/front design. All results demonstrate that our model can capture perceptual 3D characteristics while delivering high levels of creativity, visual appeal.}
    \label{fig:wireArt}

  \end{minipage}
 
  \begin{minipage}[!t]{0.49\textwidth}
    \centering
    \includegraphics[width=\textwidth]{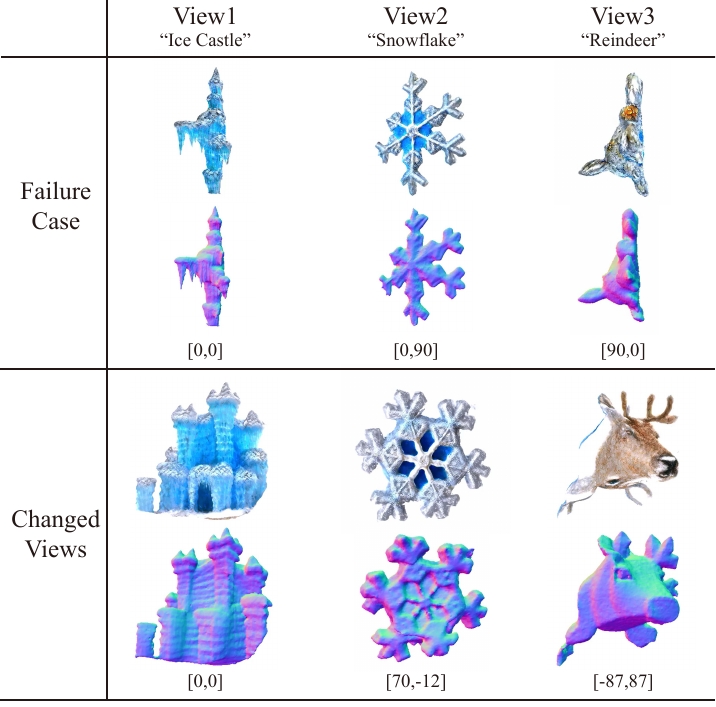}
    \caption{\textbf{Limitations of Our Results.} We present the renderings and normal maps of a failure case. Under certain semantics and observation views as inputs, the geometry we generate might collapse or distort. However, after changing the observation views, the generated results could become much more compatible.}
    \label{fig:limitation}
  \end{minipage}
  \hfill
  \raggedleft
  \begin{minipage}[!t]{0.49\textwidth}
    \includegraphics[width=\textwidth]{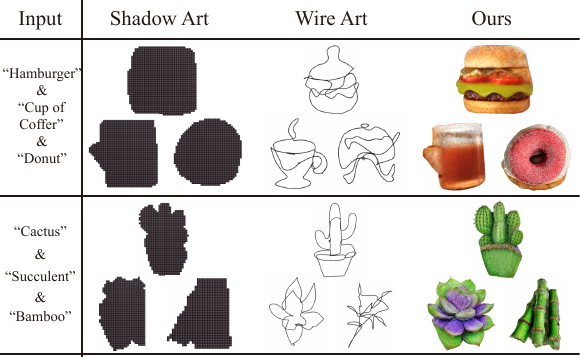}
    \caption{\textbf{Comparison with Similar Works.} We compare with Shadow Art \cite{MP09}, Wire Art \cite{Fabricable3DWireArt}. Considering that Shadow Art only accepts binary images as input, the inputs for Wire Art during comparison are RGB images we rendered, while the inputs for Shadow Art are their masks. The input views are [0,0], [0,90], [90,0]. The results illustrate that our models effectively integrate multiple semantic elements, presenting the information in a manner that is more readily perceivable to observers. }
    \label{fig:ShadowArt}
  \end{minipage}
\end{figure*}

Our method still has some limitations. Complex semantics can introduce inherent multi-view conflicts that are difficult to fully resolve. Additionally, while our strong geometric constraints effectively prevent oversimplified or flattened results, they can occasionally cause collapses or distortions. As shown in Fig.~\ref{fig:limitation}, selecting alternative input views can mitigate some of these conflicts. A promising avenue for future work is to treat observation views as optimizable parameters; this could improve semantic compatibility and allow for better integration of different shape characteristics.


\bibliographystyle{ACM-Reference-Format}
\bibliography{bibfile}

\appendix
\newpage
\section{Pseudocode}
    
\begin{algorithm}[!h] 
    \caption{Geometry Generation with LGAD}
    \begin{spacing}{1.15}
            \begin{algorithmic}[1]
                \STATE \textbf{Input}: Main views $\mathcal{V}=\{v_i\}_{i=0}^{n-1}$ and semantics. $\mathcal{Y}=\{y_i\}_{i=0}^{n-1}$
                \STATE \textbf{Choose}: multi-view ND diffusion model with noise predictor $\boldsymbol{\epsilon}_{\text {pre }}^{3 D}$
                \STATE \textbf{initialize} Tetrahedron Splatting $\theta$ as a sphere
                \WHILE{$\theta$ is not converged}
                    \STATE Sample: camera $c$ in the neighbor of $\mathcal{V}$, horizontal surround cameras $\mathcal{C}=\{c_i\}$
                    \STATE $\{w_i\} = \{1/(1-c\cdot v_i)\}$, $\{w^\prime_i\} = sort(\{w_i\})$
                    \STATE $y(c)=y_{argmax\{w^{\prime}_i\}}$
                    \STATE $\boldsymbol{g}_0 = \bm{n}_0, \bm{d}_0 = P(\theta, c)$.
                    \STATE Sample: $t, \boldsymbol{\epsilon}_{c}$.
                    \STATE copy $\boldsymbol{g}_t$ to obtain $\left\{\boldsymbol{g}_{t, c^{\prime}}\right\}_{c^{\prime} \in C}$
                    \STATE $\boldsymbol{\epsilon}_{y,c} = \boldsymbol{\epsilon}_{\text {pre }}^{3 D}\left(\left\{\boldsymbol{g}_{t, c^{\prime}}\right\}_{c^{\prime} \in C} ; t, y(c), C\right)|_c$
                    \STATE $\boldsymbol{\epsilon}_{\varnothing,c} = \boldsymbol{\epsilon}_{\text {pre }}^{3 D}\left(\left\{\boldsymbol{g}_{t, c^{\prime}}\right\}_{c^{\prime} \in C} ; t, \varnothing, C\right)|_c$
                    
    \STATE $s = s_0(w^{\prime}_0-w^{\prime}_1)/\sum_iw^{\prime}_i$
                    \STATE
                    $  \hat{\boldsymbol{\epsilon}}_{c}=s\hat{\boldsymbol{\epsilon}}_{y,c}+(1-s)\boldsymbol{\epsilon}_{\varnothing,c}$ 
                    \STATE $\nabla_{\theta} \mathcal{L}_\mathrm{LGAD}=\mathbb{E}_{t, \boldsymbol{\epsilon}, c}\left[\omega(t)\left(\hat{\boldsymbol{\epsilon}}_{c}-\boldsymbol{\epsilon}_{c}\right) \frac{\partial \boldsymbol{g}}{\partial \theta}\right]$
                    \STATE $\nabla_{\theta} \mathcal{L}_{\mathrm{SDS}}=\mathbb{E}_{t, \boldsymbol{\epsilon}, c}\left[\omega(t)\left(\boldsymbol{\epsilon}_{\phi}(\bm{n}_t; y(c), t)-\boldsymbol{\epsilon}\right) \frac{\partial \bm{n}_0}{\partial \theta}\right]$
                    \STATE 
                    $\mathcal{L}_{\text {eik }}=\sum_{\delta}\left(\left\|\nabla f_{\delta}\right\|_{2}-1\right)^{2},$
                    \STATE $\mathcal{L}_{\mathrm{nc}}=\sum_{e}\left(1-\cos \left(\boldsymbol{n}_{e_{1}}, \boldsymbol{n}_{e_{2}}\right)\right)$
                    \STATE update $\theta$ with losses above
                \ENDWHILE
                \RETURN $\theta$
            \end{algorithmic}
    \end{spacing}
    \label{alg:lgad}
\end{algorithm}

\section{Experiment Details}
\paragraph{Selection of Semantics and observation Views}
We use ChatGPT to generate semantics, requiring certain correlations between each group of semantics to enhance aesthetic effects. When selecting observation views, some results adopt fixed patterns (e.g., horizontal surround, orthographic views), while others are randomly initialized before generation, with the constraint that the angle between any two observation views is no less than 120 degrees.
\section{More Experiment Results}

\begin{figure*}[ht]
  \centering
  \includegraphics[width=0.95\linewidth]{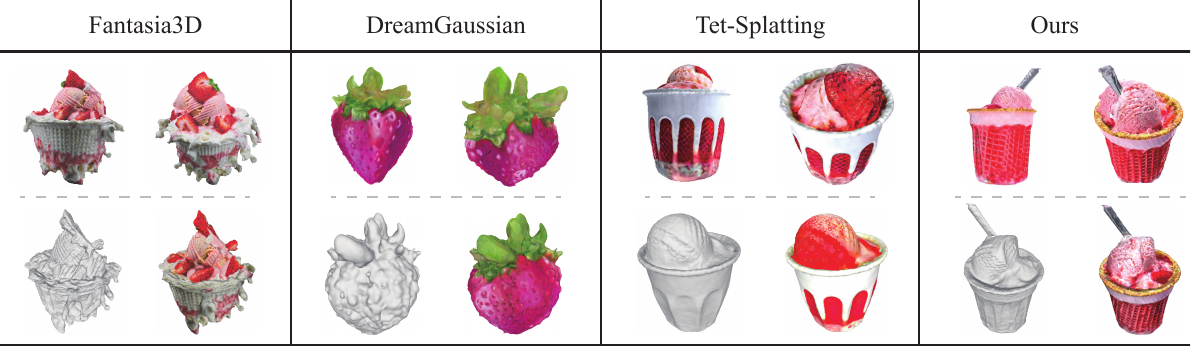}
  \caption{\textbf{Qualitative Comparison of Text-to-3D Methods on a Single Semantic Concept.} The input semantics is ``strawberry ice cream sundae in the cup''. The top row is the rendering and the bottom row is the extracted mesh with/without textures. The result of Fantasia3D~\cite{chen2023fantasia3d} contains much noise, and the result of DreamGaussian~\cite{tang2023dreamgaussian} deviates from semantics. Compared with TeT-Splatting~\cite{tetsp}, the mesh we extracted can better maintain the details in PBR.}
  \label{fig:single}
\end{figure*}

To demonstrate our generation quality, we conduct comparisons with other prior-based 3D generation works under a single semantics as input. Since the guided semantics remain invariant to view changes in such a case, we sample 360-degree cameras and keep the guidance scale constant, and utilize four orthogonal ND maps as supervision. As shown in Fig.~\ref{fig:single}, our method achieves superior geometry and rendering quality compared to established works when given identical semantic inputs. And compared with the work also adopting tetrahedral splatting representations~\cite{tetsp}, our extracted mesh better approximates physically based rendering results with finer details.

\begin{figure*}
    \includegraphics[width=0.5\linewidth]{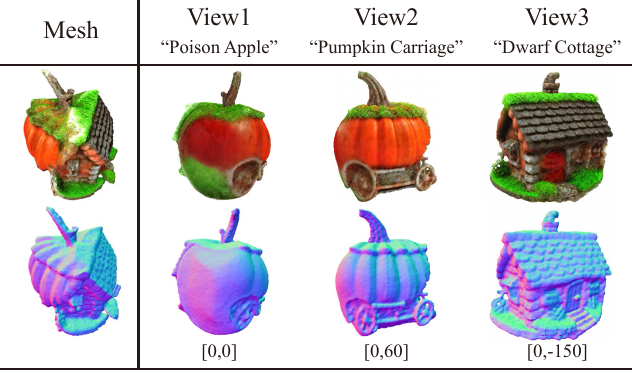}
        \caption{\textbf{Experiments with Reducing Viewing Distance.} This shape achieves generation by associating the decayed portion of an apple with a wheel's structure, which facilitates extensive geometry and rendering resource sharing.}
        \label{fig:abl_small}
\end{figure*}

We also conduct experiments with reduced distance between observation views, where shared geometry and rendering components significantly increased generation difficulty. As demonstrated in Fig.~\ref{fig:abl_small}, our method accomplishes the task even when the separation between two observation views is narrowed to 60 degrees.

\section{3D Printing Results}
We 3D-print several examples using both colored and white materials: white prints make it easy to observe geometric details, while colored prints can verify the final appearance. 
\begin{figure*}
    \centering
    \includegraphics[width=0.5\textwidth]{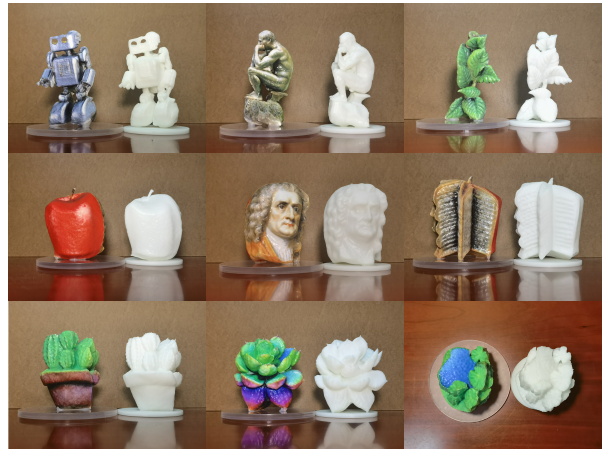}
    \caption{\textbf{3D Printing Results.} Results show that the manufactured outcomes are nearly identical to the simulations, delivering eye-catching visual effects. 
    }
    \label{fig:fab}
\end{figure*}




\end{document}